\documentclass[11pt]{article}
\usepackage{coling2020}
\usepackage{hyperref}
\usepackage{times}
\usepackage{latexsym}
\usepackage{bbding,pifont,wasysym,amssymb} % for checkmark
\usepackage{url}

\colingfinalcopy % Uncomment this line for the final submission

\usepackage[T1]{fontenc}
\usepackage[scaled=.8]{beramono}
\usepackage[scaled=.95]{helvet}

\usepackage{color,soul}

\usepackage[shortlabels]{enumitem} % customizable lists
\setitemize{noitemsep,topsep=0em} %leftmargin=1em,iciteosstemindent=-1em

\usepackage{multirow}
\usepackage{tikz}
\usepackage{tikz-dependency}
\usepackage[warn]{textcomp}
\usepackage{etoolbox}
\usepackage{xr}
\usepackage{xfrac}
\usepackage{pgfplotstable}

\usepackage{tabularx}
\usepackage{collcell}
\usepackage{booktabs}

\usepackage{nameref}
\usepackage{cleveref}

\pgfplotsset{compat=1.14}

\usepackage{wrapfig}

\definecolor{orange}{rgb}{1,0.5,0}
\definecolor{mdgreen}{rgb}{0.05,0.6,0.05}
\definecolor{Acolor}{HTML}{EC5D57} % poppy red
\definecolor{Pcolor}{HTML}{70BF41} % grass green
\definecolor{Scolor}{HTML}{51A7F9} % sky blue
\definecolor{Lcolor}{HTML}{B36AE2} % friendly purple
\definecolor{mdblue}{rgb}{0,0,0.7}
\definecolor{dkblue}{rgb}{0,0,0.5}
\definecolor{dkgray}{rgb}{0.3,0.3,0.3}
\definecolor{slate}{rgb}{0.25,0.25,0.4}
\definecolor{gray}{rgb}{0.5,0.5,0.5}
\definecolor{ltgray}{rgb}{0.7,0.7,0.7}
\definecolor{purple}{rgb}{0.7,0,1.0}
\definecolor{lavender}{rgb}{0.65,0.55,1.0}

\newcommand{\com}[1]{}

\usetikzlibrary{shapes,shapes.misc}

\newcommand{\ApplyGradient}[1]{%
  \pgfmathsetmacro{\PercentColor}{(#1-0)/7.97}%
  \pgfmathsetmacro{\PercentInverse}{ifthenelse(\PercentColor > 65, 0, 100)}%
  \edef\x{\noexpand\cellcolor{red!\PercentColor}}\x\textcolor{black!\PercentInverse}{#1}%
}
\newcolumntype{R}{>{\small\collectcell\ApplyGradient}{c}<{\endcollectcell}}

\title{Refining Implicit Argument Annotation for UCCA}

\author{
  Ruixiang Cui \and Daniel Hershcovich \\
  Department of Computer Science \\
  University of Copenhagen \\
  \quad\texttt{\{rc, dh\}@di.ku.dk} \\
}

\begin{document}

\maketitle

\begin{abstract}
Predicate-argument structure analysis is a central component in meaning representations of text. The fact that some arguments are not explicitly mentioned in a sentence gives rise to ambiguity in language understanding, and renders it difficult for machines to interpret text correctly. However, only few resources represent implicit roles for NLU, and existing studies in NLP only make coarse distinctions between categories of arguments omitted from linguistic form. This paper proposes a typology for fine-grained implicit argument annotation on top of Universal Conceptual Cognitive Annotation's foundational layer. The proposed implicit argument categorisation is driven by theories of implicit role interpretation and consists of six types: Deictic, Generic, Genre-based, Type-identifiable, Non-specific, and Iterated-set. We exemplify our design by revisiting part of the UCCA EWT corpus, providing a new dataset annotated with the refinement layer, and making a comparative analysis with other schemes.

\end{abstract}

\section{Introduction}
\blfootnote{
    \hspace{-0.65cm}  % space normally used by the marker
    This work is licensed under a Creative Commons 
    Attribution 4.0 International License.
    License details:
    \url{http://creativecommons.org/licenses/by/4.0/}.}
% Meaning Representations 
Semantic representation frameworks have been a major medium to understanding the nature of languages for NLP. Through these frameworks, researchers have been exploring linguistic phenomena such as quantification \cite{pustejovsky-etal-2019-modeling}, coreference \cite{prange-etal-2019-semantically}, and word sense  \cite{schneider-etal-2018-comprehensive}. However, most efforts were put into studying  linguistic complexity superficially, rather than the more latent, implicit omission of arguments in an event. For instance, in the sentence ``Just take the money!'', the addressee who should ``take the money'' is left out. Such omission cannot be recovered directly from the text in the way of gapping or ellipsis, but require a higher level of understanding and inference from the context. Traditional studies approach argument omission from different aspects, namely syntactically, semantically, or pragmatically. The interpretation of implicit roles varies to a great extent, from phonological deleted role during production \cite{perlmutter1968deep,mittwoch1971idioms,perez2018direct} to timecourse reference omission from a psycholinguistic aspect \cite{garrod2000contribution}. However, few studies have explored the implicit role phenomenon in NLP. 

In this paper, we propose a fine-grained cross-linguistically applicable implicit argument annotation typology as a refinement for Universal Conceptual Cognitive Annotation  \cite[UCCA]{abend2013universal} categories. The typology follows UCCA's design concept, focusing on the semantic notion of Scene rather than linguistic form phenomena. The proposed implicit argument set contains six categories: Deictic, Generic, Genre-based, Type-identifiable, Non-specific, and Iterated-set. We refine the existing UCCA relation labels and add information to them, while keeping all categories from the underlying annotation. Our studies move UCCA, a semantic representation framework, to have less syntactic definitions of implicit arguments.

Based on the proposed typology, we conduct a pilot annotation study, including revisit and refinement of the UCCA EWT dataset,\footnote{\url{https://github.com/ruixiangcui/UCCA-Refined-Implicit-EWT\_English}} and subsequently make a comparative analysis with the only other existing fine-grained implict role annotation scheme, Fine-grained Annotations of Referential Interpretation Types  \cite[FiGref]{o2019bringing}.
\begin{figure}
\centering
\begin{tikzpicture}[->,level distance=1.35cm,
  level 1/.style={sibling distance=4cm},
  level 2/.style={sibling distance=20mm},
  level 3/.style={sibling distance=20mm},
  every circle node/.append style={fill=black},
  every node/.append style={text height=1ex,text depth=0}]
  \tikzstyle{word} = [font=\rmfamily,color=black]
  \node (1_1) [circle] {}
  {
  child {node (1_2) [circle] {}
    {
    child {node (1_16) [word] {\textbf{IMP}$_\text{Non-specific}$}  edge from parent node[midway, fill=white]  {A}}
    child {node (1_8) [word] {Have}  edge from parent node[midway, fill=white]  {D}}
    child {node (1_9) [circle] {}
      {
      child {node (1_13) [word] {a}  edge from parent node[midway, fill=white]  {F}}
      child {node (1_12) [word] {real}  edge from parent [white]}
      child {node (1_14) [word] {mechanic}  edge from parent node[midway, fill=white]  {C}}
      } edge from parent node[midway, fill=white]  {A}}
    child {node (1_10) [word] {check}  edge from parent node[midway, fill=white]  {P}}
    } edge from parent node[midway, fill=white]  {H}}
  child {node (1_3) [word] {before}  edge from parent node[midway, fill=white]  {L}}
  child {node (1_4) [circle] {}
    {
    child {node (1_6) [word] {you}  edge from parent node[midway, fill=white]  {A}}
    child {node (1_7) [word] {leave}  edge from parent node[midway, fill=white]  {P}}
    child {node (1_17) [word] {\textbf{IMP}$_\text{Non-specific}$}  edge from parent node[midway, fill=white]  {A}}
    } edge from parent node[midway, fill=white]  {H}}
  };
  \draw[dashed,->] (1_2) to node [midway, fill=white] {A} (1_6);
  \draw[bend right,->] (1_2) to[out=-20, in=180] node [midway, fill=white] {D} (1_12);
\end{tikzpicture}
    \caption{Example of UCCA graph: ``Have a real mechanic check before you leave.''. Abbreviation of UCCA edge labels is explained in Table 1. The dashed line stands for \textit{Remote} edge. In this case it is a coreference "you". An \textbf{IMP} represents an \textit{Implicit} argument denoting a null-instantiated core element in its corresponding Scene. }
    \label{fig:example sentence}
\end{figure}
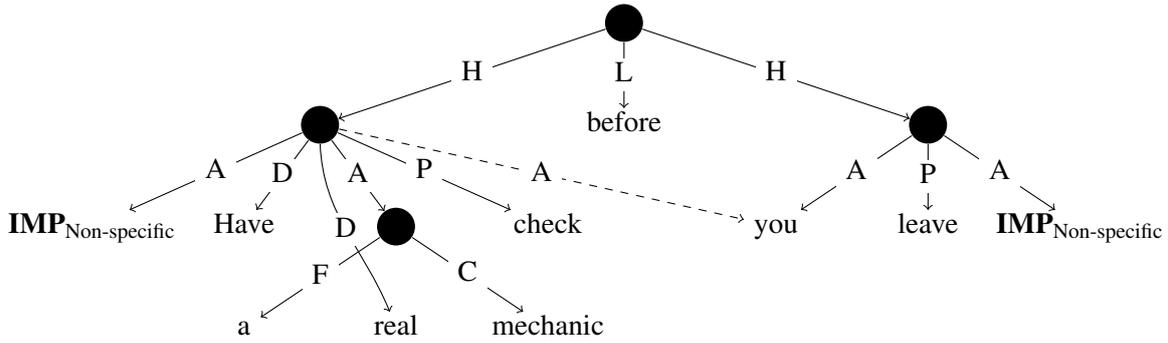

A few studies have explored the possibilities to parse implicit arguments. Both \newcite{gerber-chai-2012-semantic} and \newcite{cheng2019implicit} have developed parsers to recover implicit arguments for nominal predicates; \newcite{bender-etal-2011-parser} parses against some linguistic constructions, two of which could license implicit arguments, that is, tough adjectives and verbal gerund; \newcite{elazar2019s} focus on resolving missing numeric fused-heads, which are implicit Centers in UCCA.

Providing the most up-to-date and fine-grained annotation of implicit arguments, our studies can potentially enhance natural language understanding if supplied with an adequate parser. For example, when companies conduct satisfaction analyses through web reviews, customers often express themselves colloquially in these reviews. Examples include ``Serves bad ice cream, Joe's is better'' and ``Near a nice district, bad service and expensive.'' If these reviews are annotated with Genre-based implicit arguments, referring to the conventional omission of reviewees, algorithms can study which part of the reviews really refers to the companies rather than other entities, and make better predictions, despite the omission of subjects and non-standard language.

\section{Universal Conceptual Cognitive Annotation}

\begin{wraptable}{r}{6.5cm}
	\centering
\begin{tabular}{|ll|ll|}
\hline
Participant & A & Linker & L \\
Center & C & Connector & N \\
Adverbial & D & Process & P \\
Elaborator & E & Quantifier & Q \\
Function & F & Relator & R \\
Ground & G & State & S \\
Parellel Scene & H & Time & T\\ \hline
\end{tabular}
\caption[protect]{Legend of UCCA edge categories \cite{hershcovich2019content}}
\label{tab:Legends of UCCA edge categories}
\end{wraptable}

Universal Conceptual Cognitive Annotation is a semantic representation scheme whose design concept comes from the Basic Linguistic Theory typological framework \cite{dixon2010basica,dixon2010basicb,dixon2012basic} and Cognitive Linguistics literature \cite{croft2004cognitive}. Abstracting away from syntactic forms, it aims at representing the main semantic phenomena in text while maintaining a low learning cost and rapid annotation by non-experts \cite{abend-etal-2017-uccaapp}. Already providing datasets in English, French, and German, UCCA has demonstrated its cross-linguistic applicability in several languages and has become a popular target framework in multiple parsing tasks \cite{hershcovich2019shared,oepen-etal-2019-mrp}. Abend et al. \shortcite{abend-etal-2017-uccaapp} have also developed an open-source web-based annotation system, UCCAApp, which supports fast annotation for linguistic representations. The effectiveness and efficiency in annotating and refining UCCA have been proven by several studies \cite{shalev-etal-2019-preparing,prange-etal-2019-semantically}.

UCCA presents the meaning of a sentence with a directed acyclic graph (DAG) whose terminal nodes correspond either to surface lexical tokens or  extra units representing implicit arguments.  Non-terminal nodes correspond to semantic units that participate in some super-ordinate relation. Edges are labeled with the role of a child node related to its parent node. The basic notion in UCCA is Scene, describing a state, action, movement, or some other relation that evolves in time. Each Scene involves one main relation (Process or State), and one or more Participants, including locations, abstract entities and sub-Scenes serving as arguments. 

Furthermore, UCCA distinguishes \textit{primary} edges, appearing explicitly in one relation, from \textit{remote} edges, allowing a Scene to indicate its arguments by linking from another Scene. \textit{Primary} edges form a tree while \textit{Remote} edges allow reentrancy, forming a DAG. In some cases, an entity of importance in the interpretation of a Scene does not explicitly exist in the text. Hence, UCCA introduces the notion of \textit{Implicit} Units to represent such kind of entity. 

For instance, the sentence ``Have a real mechanic check before you leave'' in Figure \ref{fig:example sentence} contains two Scenes, evoked by ``check'' and ``leave''. The individual Scenes are annotated as follows:

\begin{enumerate}
    \item ``(\textit{You})$_A$ have$_D$[a$_F$ real$_D$ mechanic$_C$]$_A$ check$_P$ \textit{IMP}$_{1 A}$''
    \item ``You$_A$ leave$_P$ \textit{IMP}$_{2 A}$''
\end{enumerate}

``A mechanic'' is a Participant in the first Scene, while ``You'' is Participant in both Scenes, as \textit{Remote} constituting reentrancy in the first one (shown in dashed line), but explicit in the latter super-ordinate relations. In the first Scene, ``Check'' is the main Process used in the causative construction, requiring three Participants. While ``You'' refers to the customer making a request and ``a real mechanic'' is the service provider who should check something,  the object that needs to be ``checked'' is missing. Therefore, we introduce an IMP A node to symbolize it. In the second Scene, ``leave'' is the Process meaning someone moves away from a source location. Although we state our little concern for non-core elements like location in this study in \S  \ref{UCCA Implicit Argument Typology}, the case is different for the ``leaving'' Scene since the source location is vital to the understanding of the departing action.  For this reason, we add an IMP A to represent the place that is being left. 

In UCCA's foundational layer, only limited cases of implicit arguments have been annotated. The main focus is on omission licensed by certain grammatical structures,\footnote{\url{https://github.com/UniversalConceptualCognitiveAnnotation/docs/blob/master/guidelines.pdf}} whose notion is similar to  \textit{Constructional Null Instantiation} in FrameNet \cite{ruppenhofer2006framenet}. Two typical examples of such constructions are imperatives (forced omission of subjects) and passives (agent omission) in English:

\begin{enumerate}
    \item Imperative: \textit{IMP}$_A$ Do$_F$ n't$_D$ bother$_P$.
    \item Passive: [The$_F$ doctor$_C$]$_A$ has$_F$ already$_T$ been$_F$ paid$_P$ \textit{IMP}$_A$.
\end{enumerate}

Several other kinds of constructions are mentioned in UCCA foundation layer guidelines, such as infinitive clause, gerund, and \textit{thank} construction.

\begin{enumerate}[resume]
    \item Infinitive clause: Is$_F$ there$_F$ [no$_E$ other$_E$ Verizon$_C$]$_A$ \textit{IMP}$_A$ to$_F$ go$_P$ to$_R$ [around$_R$ downtown$_C$]$_A$?
    
    \item Gerund: How$_D$ addicting$_S$ \textit{IMP}$_A$ going$_P$ [to$_R$ Fitness Unlimited$_C$]$_A$ can$_FD$ be$_F$!''
    \item ``\textit{Thank}'' construction: \textit{IMP}$_A$ Thanks$_P$, John$_A$!
\end{enumerate}

The annotation of implicit arguments in UCCA's foundation layer is restricted to specific linguistic constructions, which is coarse, language-specific, incomplete, and unlike other distinctions in UCCA, is based on criteria of form rather than meaning. We are faced with the challenge of maintaining UCCA's idiosyncrasy of differentiating \textit{Remote} and \textit{Implicit} while extending its boundary to include a rather refined categorisation for implicit roles.

\section{Implicit Roles In Meaning Representations}
\subsection{Background and Motivation}
FrameNet \cite{ruppenhofer2006framenet} is a source of inspiration for UCCA and Scenes can be seen as frame evocations. FrameNet developed Fillmore \shortcite {fillmore1986pragmatically}'s notion of null-instantiation into three types---Constructional Null Instantiation (CNI), licensed by grammatical constructions, Definite Null Instantiation (DNI), equivalent to core Frame Elements mentioned previously in text or inferrable from the discourse, and Indefinite Null Instantiation (INI), an  element that is unknown and nowhere to be retrieved. 

Nevertheless, this trichotomy treats unfavourably many cases where implicit roles occur, such as Free Null Instantiation \cite[FNI;]{fillmore1993construction} and Identity of Sense Null Anaphora \cite[ISNA;]{kay2006null}. FNI is neither restricted to definite nor indefinite null arguments, and ISNA is null instantiation within noun phrases. Lyngfelt \shortcite{lyngfelt2012re} even argues that the unclear definition of FNI leads to much false categorisation---some FNIs are unspecified, some are generic, and some should be considered DNI. 

UCCA's foundational layer mainly focuses on CNI, that is, the current annotated datasets only include grammatically licensed implicit arguments. So far, only a few corpora for implicit role labelling have been proposed, such as SemEval-2010 Task 10 \cite{ruppenhofer-etal-2009-semeval}, Beyond NomBank \cite{gerber-chai-2010-beyond,gerber-chai-2012-semantic}, and Multi-sentence AMR \cite{ogorman-etal-2018-amr}. But none of them is based on a more comprehensive fine grained implicit role characterisation theory aside from FiGref, refined on three corpora mentioned above for recoverability studies.

Although FiGref is not available to the public, \newcite{o2019bringing} has counterbalanced previous studies on implicit role description and synthesized an inventory of eleven interpretation types for implicit roles distinguished by their referential behaviours. They are Script-inferrable pragmatic, Salient/recent, Deictic, Remembered Roles of Event Reference, Implicit ``Sloppy anaphora'' and Bridging, Genre-based Default, Type-identifiable, Generic (``People in General''), Cataphoric, Low-information, and Iterated Events Implicit Roles.

Recoverable implicit roles fall into the category of \textit{Remote} Participants in UCCA, which typically calls for coreference resolution \cite{prange-etal-2019-semantically}.  We discern the  eleven types of implicit roles mentioned above, recoverable or not, and extract those types where only true \textit{Implicit} arguments occur, that is, the argument cannot be explicitly recovered from text, but inference and non-specificity are allowed as they are aligned with the definition of implicit arguments of UCCA. In the following section, we will analyze these implicit role types and argue the appropriateness of the set of categories we choose. 

For the sake of operability and consistency, we only focus on core arguments in Scenes where these arguments are essential to the meaning of corresponding predicates \cite{jackendoff1992semantic,jackendoff1997twistin,goldberg1992inherent,grimshaw1993semantic}. Elements such as location, time, and manner are of little interest in this study whilst they are able to appear as foundational units or \textit{Remote} Participant in UCCA.

As UCCA distinguishes \textit{Remote} edges from \textit{Implicit} units, it is natural to take advantage of this property to account for argument recoverability. The definition of implicit arguments in UCCA, particularly for its strong emphasis on the inability of explicit recovering from text, is not strictly corresponding with the eleven implicit role types. 

\subsection{Forming UCCA Implicit Argument Typology}
\label{UCCA Implicit Argument Typology}
\begin{table*}[t]
\centering
\begin{tabular}{@{}lccc|r@{}}
O'Gorman 2019 & Definite & Indefinite & Edge Cases & UCCA's Implicit Refinement \\ \hline
Salient/recent & \checkmark &  &  & \ding{55}  \\
Remember Roles & \checkmark &  &  & \ding{55} \\
Script-inferrable & \checkmark &  &  & \ding{55} \\
Deictic &\checkmark  &  &  & Deictic \\
Cataphoric &  & \checkmark &  & \ding{55} \\
Low-information &  & \checkmark &  & Non-specific \\
Iterated Events &  & \checkmark &  & Iterated-set \\
Bridging &  &  & \checkmark & \ding{55} \\
Genre-based &  &  & \checkmark & Genre-based \\
Generic &  &  & \checkmark & Generic \\
Type-Identifiable &  &  & \checkmark & Type-Identifiable \\ 
\end{tabular}
\caption{The 11 implicit role types in O'Gorman 2019 and the set for UCCA's implicit refinement.}
\label{O'Gorman's categories and the set for UCCA's implicit argument typology}
\end{table*}

Table \ref{O'Gorman's categories and the set for UCCA's implicit argument typology} shows the comparison of the primary eleven implicit role types and UCCA's implicit argument set. Among these types of implicit roles in his inventory, four are definite implicit role constructions, viz. Salient/recent, Remember Roles, Script-inferrable, and Deictic. Only the last one of four, Deictic,  we would consider a candidate category for UCCA's \textit{Implicit} arguments. Salient/recent roles, which can be directly found in the recent prior discourse, is the quintessential type of DNI, and they can be easily replaced by pronouns. Remember Roles and Script-inferrable roles, however, require a  cognitive  and reasoning process, as the referents can be understood through a common ground in the text shared between the speaker and the addressee, or are reflecting a different facet of the same or a subordinate event. Deictic roles, albeit explicit reference to the speaker or addressee, is an extra-linguistic and cannot be annotated as \textit{Remote} Participant, since we are unable to retrieve them explicitly in the text. Therefore, we incorporate it in our set of \textit{Implicit} arguments categories. 

Three out of eleven implicit roles are marked as clearly indefinite arguments, namely Cataphoric, Low-information Arbitrary role, and Iterated Events Implicit Roles. Cataphoric, which Bhatia et al. \shortcite{bhatia-etal-2014-unified} define as ``pragmatically specific indefinite'', is the only type we do not include in UCCA's typology since it relies heavily on the interpretation of the discourse whether it will be referred to again. We would  annotate it as \textit{Remote} Participant if the role is mentioned in a later text, or Non-specific type if not so as not to complicate the reasoning process. 

The other four, Bridging implicit roles, Genre-based Default, Generic and Type Identifiable, are regarded as edge cases. Once again, we will only admit the latter three in our typology. As far as we are concerned, bridging in ellipsis situations might not refer to the same referent conceptually. Nonetheless, it can be clearly resolved in the text. Therefore, it will be annotated as a \textit{Remote} edge in UCCA. 

We will focus on referents that do not appear anywhere in the text. Therefore, we follow the philosophy of UCCA and propose six categories of \textit{implicit argument}, that is, Deictic, Generic, Genre-based, Type-identifiable, Non-specific and Iterated-set. In the next section, we will present and exemplify each one of them.

\subsection{Categorisation Set for  UCCA Implicit Arguments}
\label{Categorisation Set for  UCCA Implicit Arguments}
\setenumerate[1]{label=(\arabic*)}

\subsubsection{Deictic}
Deictic implicit arguments specifically refer to the speaker or the addressee in a sentence. In example 1, the second-person subject is exhorted to take a certain action, and such imperative construction allows the subject not to  appear in the text explicitly. Shown in example 2, Deictic can also occur with certain interjections, where the subject as the speaker is habitually implicit.
\begin{enumerate}[label={(\arabic*)}]
    \item Just$_D$ ask$_P$ them$_A$ exactly$_E$ what$_C$ [they$_A$ want$_P$ (\textit{what})$_A$]$_E$ \textit{IMP}$_{Deictic}$. 
    \item {[}Thank you]$_P$ guys$_{G/A}$ \textit{IMP}$_{Deictic}$.
\end{enumerate}

 It should be mentioned that only in certain languages is imperative likely to induce implicit arguments. In Romance languages such as Spanish,  deictic information tends to be encoded morphologically due to person agreement \cite{ingram1971towardv}.

\subsubsection{Generic}
Generic implicit arguments denote ``people in general'' \cite{lambrecht2005definite}. In example 3, the agent who ``understands how this place has survived the earthquake'' is not explicitly mentioned in the text, but it can be understood as it is the set of people in general.  Example 4 can be construed as a gerund construction. ``I'' recommend taking a certain action. While the patient would not be specific, it conveys the message that ``people in general'' should follow such advice.

\begin{enumerate}[resume]
    \item It$_F$ 's$_F$ impossible$_D$ to$_F$ understand$_P$ [how$_C$ [[this$_E$ place$_C$]$_A$ has$_F$ survived$_P$ [the$_F$ earthquake$_C$]$_A$]$_E$]$_A$ \textit{IMP}$_{Generic}$.
    \item I$_A$ would$_F$ recommend$_P$ [not$_D$ using$_P$ [this$_E$ company$_C$]$_A$ \textit{IMP}$_{Generic}$]$_A$. 
\end{enumerate}

\subsubsection{Genre-based}
Ruppenhofer and Michaelis \shortcite{ruppenhofer2010constructional} found certain text genres, namely instructional imperatives, \textit{labelese}, diary style, match report, and judgment-expressing quotative verbs, are closely linked with conventional omission. UCCA EWT corpus is based on online reviews of businesses and services by individuals. The review genre is so prominent acoss all dataset that it forms a pattern where the reviewers do not bother to mention the reviewees explicitly. In example 5 and 6, the review genre licenses the omission of the deliverer of the action ``deliver'' and server of the action ``serve'', as they refer to the reviewees by default, because in both contexts it is the restaurants that are being reviewed. 

\begin{enumerate}[resume]
    \item Delivery$_P$ is$_F$ [lightning$_E$ fast$_C$]$_D$ \textit{IMP}$_{Genre-based}$ \textit{IMP}$_{Non-specific}$!
    
    \item {[}Great$_D$ service$_P$ \textit{IMP$_{Genre-based}$} \textit{IMP$_{Generic}$}]$_H$ and$_L$ [awesome$_S$ prices$_A$]$_H$. 
\end{enumerate}

\subsubsection{Type-identifiable}
\label{Type-identifiable}
There exist some predicates allowing listeners to naturally think they ``know'' the omitted referents because of their high predictability. In example 7, the vague referent of ``eat'' can be understood from an inherent understanding of the listeners as ``some kind of food``. In example 8, the thing that ``I drive'' is not mentioned. Instead, it comes from common sense that the referent should be a kind of vehicle. Whatever kind it is, the lack of explicit mention barely affects the understanding of the text.
\begin{enumerate}[resume]
    \item It$_A$ is$_F$ my$_A$ favourite$_S$ [place$_C$ [(\textit{my})$_A$ to$_F$ eat$_P$ \textit{IMP}$_{Type-identifiable}$]$_E$]$_A$.
    \item I$_A$ 'll$_F$ drive$_P$ [an$_Q$ hour$_C$]$_T$ [just$_E$ for$_R$ [their$_S$ (volcano)$_A$]$_E$ volcano$_C$]$_A$ \textit{IMP}$_{Type-identifiable}$.
\end{enumerate}

\subsubsection{Non-specific}
Non-specific implicit arguments refer to the kind of referents that cannot be inferred or understood at all. Such required information absent from the context attributes to the low interpretability of the implicit arguments, leaving them non-specific. As in example 9 and 10, it is impossible to infer what is ``delivered'' or who ``charged me``  neither from common knowledge nor given context. Such kind of implicit arguments are commonly found in nominalization and passive because there are high possibilities that not all agents/patients are always mentioned despite the fact that they might be core frame elements.
\begin{enumerate}[resume]
    \item There$_F$ is$_F$ no$_D$ delivery$_P$ \textit{IMP}$_{Genre-based}$ \textit{IMP}$_{Non-specific}$.
    \item I$_A$ don$_F$ 't$_D$ think$_P$ [I$_A$ have$_F$ ever$_T$ been$_F$ charged$_P$ before$_T$ \textit{IMP}$_{Non-specific}$]$_A$. 
\end{enumerate}

\subsubsection{Iterated-set}
\textit{Iterated-set} implicit arguments refer to a heterogeneous set of entities when the predicates are often an action that happens repeatedly, either iteratively or generically \cite{goldberg2001patient}. For example, in sentence 11, the predicate ``wait'' implies high repetition, and the set of patients of ``what/who I am waiting for'' is so general that it does not hold any meaning beyond the context. As in example 12, the action `` steal'' designates a Scene where anything could be stolen, but ``I'' do not and will never steal. Unlike \S \ref{Type-identifiable} Type-identifiable referring to a specific type of referents, the set of ``things'' in Iterated-set points to a vague set of entities to fill a role in a more functional way.
\begin{enumerate}[resume]
    \item I$_A$ never$_T$ wait$_P$ [in$_R$ the$_F$ waiting$_E$ room$_C$]$_A$ [[more$_C$ [than$_R$ two$_C$]$_C$]$_Q$ minutes$_C$]$_T$ \textit{IMP}$_{Iterated-set}$.
    \item I$_A$ don$_F$ 't$_D$ steal$_P$ \textit{IMP}$_{Iterated-set}$. 
\end{enumerate}

\subsection{Inherent Ambiguity and  ``Continuum'' of Coreness }
\subsubsection{Category Priority}
There are a few cases when it is difficult to choose between two categories. Since UCCA does not aspire to annotate all possible interpretations \cite{abend2013universal}, the annotator should make a best guess and choose one option. The first one is between Deictic and Generic, shown in example 13. The second one is between Genre-based and Non-specific, as in example 14. To keep the annotation consistent and maintain as much information as possible, we always choose Deictic over Generic, and Genre-based over Non-specific if available. 
\begin{enumerate}[resume]
    \item {[}The$_F$ experience$_C$]$_A$ [with$_R$ every$_Q$ department$_C$]$_A$ has$_F$ been$_F$ great$_D$ \textit{IMP}$_{Deictic}$.
    \item I$_A$ will$_F$ definitely$_D$ refer$_P$ [[my$_A$ friends$_{A/S}$]$_C$ and$_N$ [(my)$_A$ family$_{A/S}$]$_C$]$_A$ \textit{IMP}$_{Genre-Based}$.
\end{enumerate}
\subsubsection{Nominalization As Occupation}
It is a judgment call whether the patient of Process instantiated by a profession should be annotated at all. Even so it remains debatable which category such kind of implicit argument belongs to.  In the current version of corpus, we will always annotate it as Type-identifiable. As in example 15, the patients of whom has been taught is unclear but neither require clarification. The Scene of teacher/teaching is annotated with a Type-identifiable implicit argument denoting a type of people recieving education. 

\begin{enumerate}[resume]
    \item They$_A$ are$_F$ [very$_E$ good$_C$]$_D$ teachers$_{A/P}$ \textit{IMP}$_{Type-identifiable}$.
\end{enumerate}

\section{Refined Implicit Corpus}
In furtherance of investigating the characteristics of UCCA's implicit arguments, we piloted a study to revisit and refine part of English Web Treebank\footnote{\url{https://catalog.ldc.upenn.edu/LDC2012T13}} annotated with the UCCA foundational layer.\footnote{\url{https://github.com/UniversalConceptualCognitiveAnnotation/UCCA_English-EWT}} 200 passages were randomly selected for experiment from the total 723 passages comprising the UCCA EWT dataset.

We use UCCAApp to carry out annotation. The process is divided into two stages. Firstly, we create passage-level review tasks to check the existing annotation whether they contain implicit arguments, and add missing arguments if necessary. Secondly, we split passages into sentences, create tasks with refinement layer and then annotate with corresponding fine-grained implicit categories. Since all the annotation works were undertaken by one single annotator, the dataset preferably serves as a demonstration of concept, and thus further measurement of inter-annotator agreement would be desired to establish a sounder dataset.

\subsection{Revisiting Orignial EWT UCCA Dataset}

The original implicit argument annotation in EWT UCCA corpus is restricted only to put concern on constructional null instantiations, and when a unit lacks a Center or a Process/State, which is out of the scope of this study. We only regard implicit argument whose category is Participant in UCCA as valid implicit in this research. Therefore, it is necessary to modify or add missing implicit arguments in the dataset.
Considering the original corpus was annotated on passage-level, whereas our new dataset will be done on sentence-level, \textit{Remote} edges across sentences will be treated by adding a new \textit{Implicit} node under its origin Scene. 

\begin{table}[t]
\centering
\begin{tabular} {l|l|l|l|l|l}
& \# Passages & \# Passages  & \# Sentences & \# Sentences & \# Implicit  \\
&& w/ Implicit && w/ Implicit & (Valid)  \\ \hline
Original & 200 (out of 723) & 103 & 306 & 111 & 153 (98) \\
Refined  & 200 & 116 & 393 & 221 & \textbf{415 (385)}
\end{tabular}%
\caption{Statistics of the UCCA EWT dataset sampled passages before and after reviewing. The additional implicit arguments result from both reviewing original UCCA EWT and conducting new annotation according to out fine-grained typology. Implicit (Valid) denotes implicit argument whose role is Participant in UCCA.}
\label{Statistics of original UCCA EWT dataset before and after reviewing }
\end{table}

\begin{figure}[ht]
    \centering
    \includegraphics[width=0.8\textwidth]{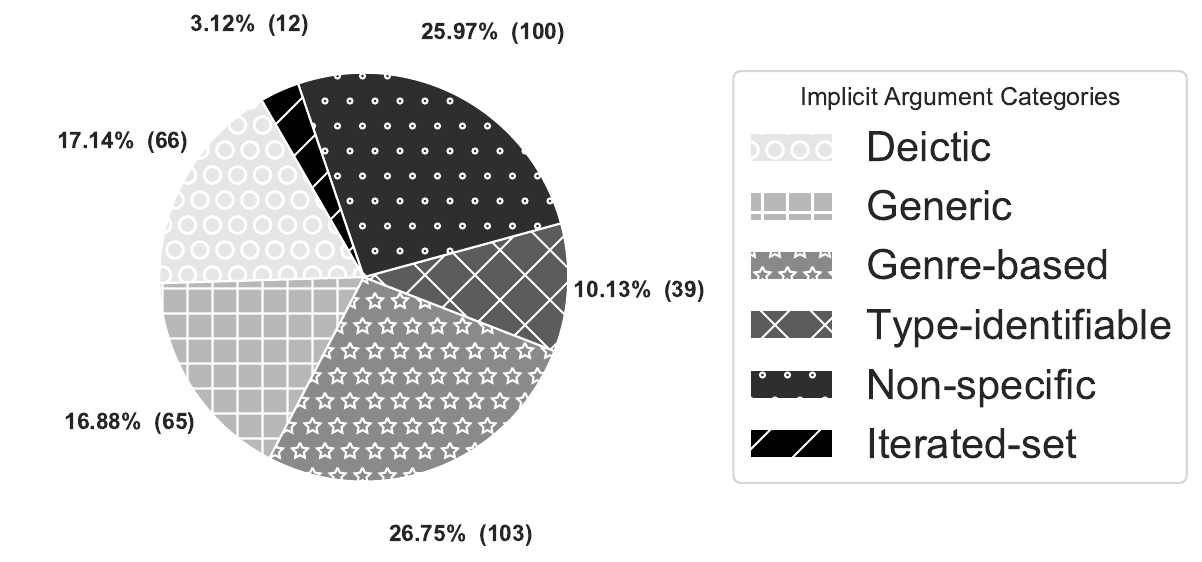}
    \caption{Statistics of pilot UCCA implicit dataset.}
    \label{fig:Statistics of Pilot UCCA Dataset}
\end{figure}

Table \ref{Statistics of original UCCA EWT dataset before and after reviewing } shows the statistics before and after reviewing and refining according to the new UCCA Implicit Argument Typology. It can be seen that in the refined dataset, 116 out of 200 passages contain implicit arguments, 13 more passages than the original dataset, in which only 103 passages contain implicit arguments. Yielding an increase of 255\%,  the review process added 250 more valid implicit arguments in the corpus. 

\subsection{Statistics}
Tokenized and split according to the Universal Dependencies English Web Treebank \cite{silveira14gold,hershcovich2019content}, this pilot corpus consists of 3702 tokens, 1411 nodes, and 4759 edges over 393 sentences. In total, 385 valid implicit arguments are found and annotated on 221 sentences. Figure \ref{fig:Statistics of Pilot UCCA Dataset} demonstrates each implicit argument category with its corresponding number in the pilot refined implicit corpus, and illustrates the percentage of each implicit category in the dataset. 
One can see that Genre-based and Non-specific are the two most frequent categories, both of which have more than 100 instances in the dataset, making up approximately 52\% combined. They are followed by Generic and Deictic, and each occupies about 17\%. Type-identifiable comes penultimate with 39 instances, while Iterated-set is the least frequent type, which merely has 12 instances, accounting for 3.12\% of the whole corpus.

\subsection{Comparisons to FiGref Annotation}
\begin{table*}[th]
\resizebox{\textwidth}{!}{%
\begin{tabular}{l|llll|ll|lll}
 & Type-Identifiable & Deictic & Generic & Non-specific & Genre-based & Iterated-set & Script-inferrable & Other & Invalid \\ \hline
Ours & 9.79\% & 17.29\% & 17.29\% & 25.93\% & 25.93\% & 2.59\% & \textbackslash{} & \textbackslash{} & \textbackslash{} \\
FiGref & 7\% & 5\% & 10\% & 4\% & \textbackslash{} & \textbackslash{} & 9\% & 12\% & 53\% 
\end{tabular}%
}
\caption{Relative frequency of annotated implicit types in UCCA's refinement layer and FiGref's annoation for non-recoverable roles in Multi-sentence AMR. The first four types are shared by both annotation corpora. The following two are exclusive to UCCA's refinement layer. The last three are additonally introduced into FiGref's set of interpretation types.}
\label{Relative frequency of annotated implicit types in UCCA and FiGref}
\end{table*}

FiGref is annotated over Multi-sentence AMR training data, SemEval-2010 Task 10 training data, and Beyond NomBank. It contains 856 implicit roles classified into 14 types, which includes all 11 proposed interpretation types except Genre-based, and four more kinds of invalid roles to account for those implicit roles of low importance or deal with tricky occasions (Local-mention, Contextually-invalid, Non-predictive and Low-importance). However, FiGref has a relatively low Cohen's $\kappa$ score \cite{cohen1960coefficient} of 55.2. With 14 types to distinguish, FiGref has a relatively high annotation complexity and ambiguity.

Comparatively, we do not annotate genuinely invalid implicit role such as these four low importance kinds, since we consider they are simply not core elements of event. As UCCA differs \textit{Implicit} units from \textit{Remote} edges, it naturally reduces the annotation complexity. Therefore, we do not have to distinguish and annotate these five types in O'Gorman 2019, viz.: Saient/recent, Remember Roles, Script-inferrable, Cataphoric and Bridging. We consdier them belong to \textit{Remote} in UCCA.
We only need to attend to essential and unrecoverable implicit arguments within the set of six types we propose in \S \ref{Categorisation Set for  UCCA Implicit Arguments}. 

Owing to the distinct annotation design and lack of statistics provided by FiGref, it is difficult to perform a comparative quantitative study between UCCA's implicit refinement layer and FiGref. However, we are able to look into the relative frequency of annotated implicit types in UCCA and FiGref's annotation for non-recoverable roles in Multi-sentence AMR shown in Table \ref{Relative frequency of annotated implicit types in UCCA and FiGref}. 

The distribution distinction can be possibly explained by the different domains of the corpora and their annotation methodology. We keep Genre-based type to account for the particular ``review'' genre in EWT dataset. Among the three types FiGref has introduced, We can see that invalid roles dominate the FiGref annotation with 53\%. This is because a large amount of non-important interpretations of null-instantiation are taken into account in FiGref, whereas the implicit refinement annotation designed for UCCA is limited to essential important implicit arguments so as to lower annotation complexity and ambiguity.

\section{Conclusion and Future Work}
We proposed a novel typology for different implicit arguments in UCCA, which allows annotation with a relatively low cognitive load. Then we reviewed and refined part of the existing UCCA English Web Treebank dataset and piloted annotation of our guidelines with a refinement layer of fine-grained implicit arguments. It is currently the only published dataset with this kind of information.

Our work addresses a deep linguistic problem in implicit role interpretation, impossible to deal with current machine learning approaches, and provides an example for conducting research about it. It has the potential to benefit natural language understanding, especially in genre-specific context, e.g., social media and customer reviews.
In this work we only use the existing definitions from \newcite{o2019bringing}, testing their compatiblity with the UCCA guidelines and their applicability to an UCCA annotated corpus. In future work we would also like to reason about and possibly redesign the criteria based on semantic and pragmatic considerations \cite{lyngfelt2012re}. We will also consider the usefulness of the refined scheme for practical NLP tasks, and possibly discard some of the categories, such as Iterated-set, which is the least frequent category in our corpus.

It is anticipated that our study will inspire tailored design of implicit role annotation in other meaning representation frameworks. Downstream tasks such as coreference resolution and human-robot interaction are also likely to benefit from reducing ambiguity and increase contextual understanding by explicitly modelling implicit arguments.

While the pilot design is promising, existing parsers that tackle implicit arguments only attend to certain aspect of linguistic phenomena. Therefore, for a rather comprehensive implicit argument annotation the like of UCCA, it is crucial to develop a parser that can emit implicit nodes dynamically and label them with fair accuracy.

Finally, work on reviewing the corpus by a second annotator is underway to validate the annotation quality by providing an inter-annotator agreement measurement. Future work will also expand the corpus and extend it to multiple languages.

\section*{Acknowledgements}
We would like to thank the anonymous reviewers, Omri Abend and Dotan Dvir for their helpful feedback.

\bibliographystyle{coling}
\bibliography{references}

\end{document}